\documentclass[12pt]{article}

\pdfoutput=1
\usepackage{amsmath}
\usepackage{graphicx}
\usepackage{enumerate}
\usepackage{natbib}
\usepackage{url} 
\usepackage{amsthm}
\usepackage[normalem]{ulem}

\RequirePackage{amsthm,amsmath,amsfonts,amssymb}
\usepackage[colorlinks, linkcolor=blue, citecolor=blue, urlcolor=blue]{hyperref}

\usepackage{tikz-cd, pgfplots, tikz}
\usepackage{color}
\usepackage{mathrsfs}
\usepackage{url}
\usepackage{caption}
\usepackage{mathtools}
\usepackage{lipsum}
\usepackage{amssymb}
\usepackage{multirow}
\usepackage{float}
\usepackage{algorithmicx}
\usepackage[ruled,vlined]{algorithm2e}
\usepackage{bbm}
\usepackage{bm}
\usepackage{booktabs}
\usepackage{stfloats}
\usepackage{enumitem}
\usepackage{xr}
\usepackage{subfig}
\usepackage[doublespacing]{setspace}
\usepackage{enumerate}
\usepackage{enumitem}
\usepackage{xr}
\usepackage{authblk}

\usepackage[utf8]{inputenc}
\usepackage[ruled,vlined]{algorithm2e}

\DeclareMathOperator*{\argmin}{argmin}

\DeclareMathOperator*{\Var}{Var}

\newcommand{\indep}{\perp\hspace*{-6.2pt}\perp}

\newcommand{\FCCov}{\text{FCCov}}
\newcommand{\FC}{\text{FC}}

\graphicspath{{./}} 

\newcommand{\blind}{0}

\theoremstyle{plain}

\newtheorem{theorem}{Theorem}[section]
\newtheorem{lemma}[theorem]{Lemma}

\newtheorem{assumption}{Assumption} 
\newtheorem*{assumption*}{Assumption} 
\newtheorem{remark}[theorem]{Remark}
\newtheorem{definition}[theorem]{Definition}

\newcommand{\N}{\mathbb{N}}

\newcommand{\R}{\mathbb{R}}
\newcommand{\F}{{\cal F}}

\newcommand{\mA}{\mathcal{A}}

\newcommand{\mC}{\mathcal{C}}
\newcommand{\mD}{\mathcal{D}}

\newcommand{\mF}{\mathcal{F}}
\newcommand{\mG}{\mathcal{G}}
\newcommand{\mH}{\mathcal{H}}

\newcommand{\mL}{\mathcal{L}}

\newcommand{\mS}{\mathcal{S}}

\newcommand{\mV}{\mathcal{V}}
\newcommand{\mM}{\mathcal{M}}
\newcommand{\mO}{\mathcal{O}}

\newcommand{\mY}{\mathcal{Y}}

\newcommand{\mf}{\boldsymbol{f}}

\def\blind{1}

\begin{document}
\date{}

\def\spacingset#1{\renewcommand{\baselinestretch}%
{#1}\small\normalsize} \spacingset{1}


\if1\blind
{
  \title{\bf Fr\'echet Cumulative Covariance Net for Deep Nonlinear Sufficient Dimension Reduction with Random Objects}
    \author[1]{BY Hang Yuan}
    \author[2]{Christina Dan Wang}
    \author[3]{Zhou Yu}

    \affil[1]{\small School of Statistics, East China Normal University, \href{mailto:52274404018@stu.ecnu.edu.cn}{52274404018@stu.ecnu.edu.cn}}
    \affil[2]{\small New York University Shanghai, \href{mailto:christina.wang@nyu.edu}{christina.wang@nyu.edu}}
    \affil[3]{\small School of Statistics, East China Normal University, \href{mailto:zyu@stat.ecnu.edu.cn}{zyu@stat.ecnu.edu.cn}}
  \maketitle
} \fi

\if0\blind
{
  \bigskip
  \bigskip
  \bigskip
  \begin{center}
    {\LARGE\bf Title}
\end{center}
  \medskip
} \fi

\bigskip
\begin{abstract}
Nonlinear sufficient dimension reduction\citep{libing_generalSDR}, which constructs nonlinear low-dimensional representations to summarize essential features of high-dimensional data, is an important branch of representation learning.  
However, most existing methods are not applicable when the response variables are complex non-Euclidean random objects, which are frequently encountered in many recent statistical applications.
In this paper, we introduce a new statistical dependence measure termed Fr\'echet Cumulative Covariance (FCCov) and develop a novel nonlinear SDR framework based on FCCov. Our approach is not only applicable to complex non-Euclidean data, but also exhibits robustness against outliers.
We further incorporate Feedforward Neural Networks (FNNs) and Convolutional Neural Networks (CNNs) to estimate nonlinear sufficient directions in the sample level. 
Theoretically, we prove that our method with squared Frobenius norm regularization achieves unbiasedness at the $\sigma$-field level. Furthermore, we establish non-asymptotic convergence rates for our estimators based on FNNs and ResNet-type CNNs, which match the minimax rate of nonparametric regression up to logarithmic factors.
Intensive simulation studies verify the performance of our methods in both Euclidean and non-Euclidean settings.
We apply our method to facial expression recognition datasets and the results underscore more realistic and broader applicability of our proposal.

\end{abstract}

\noindent%
\vfill

\newpage
\spacingset{1.9} 

\section{Introduction}\label{sec:intro}
Sufficient dimension reduction (SDR) serves as a prominent framework in supervised learning, aiming to find an intrinsic low-dimensional representations of high-dimensional data \( X \) while preserving essential information related to the responses \( Y \). Since the introduction of sliced inverse regression (SIR) \citep{li1991sliced} for the purpose of linear sufficient dimension reduction, various linear SDR methods have been developed. These include sliced average variance estimates (SAVE) \citep{cook1991discussion}, contour regression (CR) \citep{li2005contour}, directional regression (DR) \citep{li2007directional}, and many others. However, linear functions may not adequately represent high-dimensional complex data, such as images and natural languages, due to the inherent nonlinearity of the data. 
To address these limitations,
\citet{cook2007fisher} further proposed nonlinear sufficient dimension reduction.
The integration of the kernel trick 
with SDR concepts inspires a series of important developments towards nonlinear 
SDR \citep{wu2008,hsing2009rkhs, yeh2009, li2011principal,libing_generalSDR,ying2022frechet}.
However, kernel approaches are computationally intractable, scaling as \(\mathcal{O}(n^3)\) for a dataset of sample size \(n\).

As data collection methods continue to diversify, statisticians face increasingly complex data types. Non-Euclidean data, as a significant and widely applicable category, has gradually become a focal point of research in the fields of machine learning and artificial intelligence.
Examples of non-Euclidean data include, but are not limited to, shapes,  
graphs, 
symmetric positive definite (SPD) matrices,
Riemannian manifold structures,
and random densities. The seminal work of \citet{petersen2016functional} introduced a general Fr\'echet regression framework for data with Euclidean predictors and non-Euclidean responses. \citet{qiu2024random} further leveraged random forests to develop a weighted local nonparametric Fr\'echet regression method. Despite these advancements, these approaches share common limitations. Firstly, they are incapable of handling structured predictors, such as images and natural languages. Secondly, their performance is prone to degradation in the presence of high-dimensional predictor variables, underscoring the need for sufficient dimension reduction methods designed for non-Euclidean and complex data structures.

Our motivating example is the facial expression recognition study based on the the JAFFE (Japanese Female Facial Expression) dataset. This dataset comprises $213$ images depicting various facial expressions from 10 distinct Japanese female subjects. Each image has a resolution of $256 \times 256$ pixels. Each subject was instructed to perform seven facial expressions, including six basic emotions and a neutral expression. These images were then annotated with average semantic ratings for each basic emotion by 60 annotators. The ratings range from 1 to 5, where 5 signifies the highest intensity of emotion and 1 signifies the lowest.
Subsequently, the average scores for each emotion were calculated for each expression image, enabling the determination of the emotional distribution for each image. Then the label corresponding to each $256\times 256$ facial expression image is a  6-dimensional vectors representing probabilistic distributions of six emotions: happiness, sadness, surprise, anger, disgust, and fear. Statistical analysis of metric space valued responses with high-dimensional structured predictors is a difficult and challenging task. 

\begin{figure}[h]
        \centering
        \subfloat{
        \includegraphics[width=0.4\textwidth]{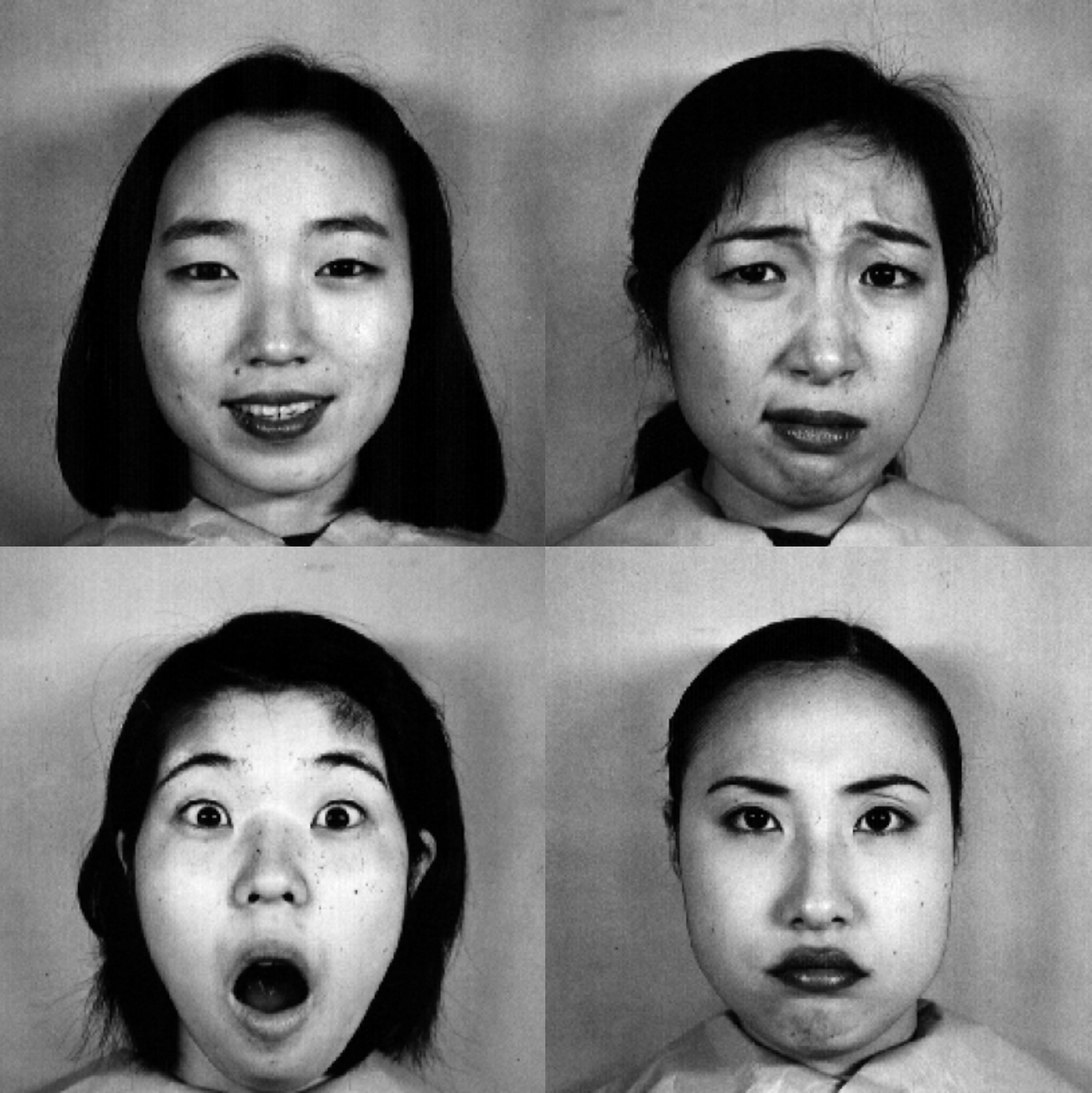}
        }
        \subfloat{
        \includegraphics[width=0.43\textwidth]{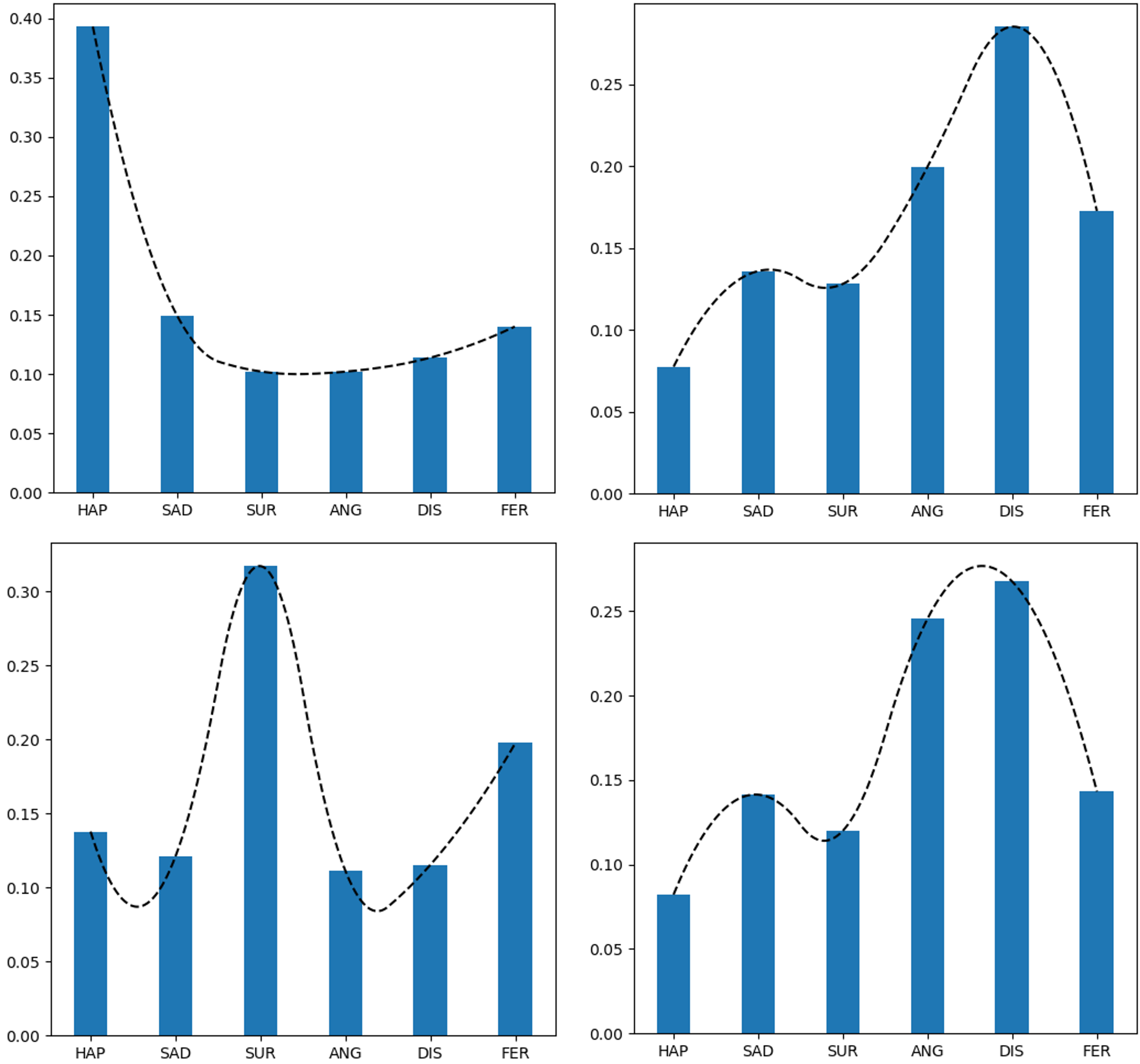}
        }
        \caption{Typical facial expressions as predictors with emotion distributions as responses.}
        \label{RAF_ML}
    \end{figure}


To this end, we in this article investigate nonlinear SDR for complex metric space valued \( Y \) and high dimensional Euclidean predictors \( X \), formalized as  
\begin{equation}\label{nonlinear_setting} 
  Y \indep X \mid \mf(X),
\end{equation}  
where \(\mf(\cdot): \mathbb{R}^p \rightarrow \mathbb{R}^d\) denotes an unknown vector-valued function of \( X \). 

Owing to the representational power and high computational efficiency, deep neural networks achieved great success in various learning tasks.
Recently, a growing number of SDR methods utilizing deep neural networks have been proposed. \citet{huang2024deep} utilized deep neural networks based on distance covariance for characterizing
nonlinear SDR.
\citet{chen2024deep} developed a nonlinear SDR framework using the generalized martingale difference divergence (GMDD) loss, optimized through neural network architectures. They also proposed two alternative optimization strategies: successive stepwise optimization and single-step optimization with Frobenius norm regularization. However, both strategies in \citet{chen2024deep} exhibit limitations. The stepwise approach is computationally inefficient due to the need for iterative updates of functional components, while the Frobenius norm regularization introduces nonsmoothness into the loss landscape, complicating gradient-based optimization.
Current deep SDR methods are primarily designed for Euclidean data and struggle to generalize to non-Euclidean domains.
Moreover, the existing theoretical investigations of deep SDR methods are insufficient for handling structured predictors, such as images structures in the MNIST data and JAFFE data.

To characterize the conditional independence between non-Euclidean random objects valued responses and Euclidean predictors, we first introduce a novel measure, termed Fr\'echet Cumulative Covariance (FCCov). By leveraging FCCov, we reformulate nonlinear SDR into a constrained optimization framework. We further develop a computationally efficient deep SDR framework that incorporates both deep fully-connected networks and ResNet-type convolutional neural networks \citep{oono2019approximation} to capture nonlinear relationships in sufficient predictors. The inclusion of ResNet architectures \citep{he2016deep} is justified by their theoretical approximation guarantees and empirical success in modeling structured data, such as images or text. Extensive simulations and real applications validate the performance of our proposal. 



The major contributions of our proposed method are summarized as follows.
\begin{itemize}
    \item 
    We develop a novel statistical metric, termed FCCov, to quantify conditional mean independence for non-Euclidean response objects and derive a corresponding fast computation algorithm. The new dependence measure is of independent interest.

    
    \item We further propose a nonlinear SDR method based on FCCov that is capable of addressing scenarios where responses are complex random objects in a metric space and predictors reside in Euclidean spaces. Theoretically, we demonstrate that our method is able to achieve unbiasedness at the $\sigma$-field level.
    
    \item Inheriting the appealing property of cumulative covariance \citep{zhou2020model, li2023testing}, our proposal with indicator functions involved demonstrates robustness to outliers. The robustness of our proposal has been substantiated through comprehensive comparisons with existing methods \citep{chen2024deep,huang2024deep}.

    \item  In contrast to \citet{chen2024deep}, which employs the Frobenius norm, we propose substituting it with squared Frobenius norm regularization, implemented via a Lagrange multiplier to reduce computational complexity.  We prove that minimizing FCCov loss equipped with squared Frobenius norm regularization can achieve unbiasedness of nonlinear SDR at the $\sigma$-field level. 
    The one-step optimization scheme with smoothed regularizers, achieves both numerical stability and computational efficiency.

    \item Last but not least, we establish the nonasymptotic convergence rate for our proposed nonlinear SDR estimators based on both
    deep fully-connected and ResNet-type convolutional neural networks.
    The convergence rate matches the minimax rate up to logarithmic terms. Existing theoretical studies \citep{chen2024deep, huang2024deep} did not consider CNNs-based estimators. Our theoretical investigations fill this gap. 
    
    

    
\end{itemize}

We organized the rest of this paper as follows. In Section \ref{sec:preliminary}, we present the necessary notations and provide a brief introduction to neural networks. In Section \ref{sec:measure}, we develop a new measure FCCov of conditional mean independence and derive a corresponding fast computation algorithm.
In Section \ref{sec:method}, we introduce the FCCov-Net, which minimizes the FCCov type loss function facilitated by neural networks, for the purpose of nonlinear SDR. Section \ref{sec:theory} establishes the nonasymptotic error bound of the proposed estimator.
In Sections \ref{sec:simulation} and \ref{sec:real_data}, we examine the proposed method through extensive numerical experiments and real data applications, respectively. Section \ref{sec:summary} discusses some future directions.

\section{Preliminaries}\label{sec:preliminary}
\subsection{Notation}
When $X$ represents a vector, we denote its components using lowercase letters $x_i$.
Consider a category of scalar-valued functions denoted by $\mF_n$, which varies with the sample size $n$. 
Unless otherwise specified, we assume that any function $f$ has an input dimensionality of $p$. For matrix inputs, such as images, the matrix is flattened into a vector, where the dimension $p$ corresponds to the dimension of the resulting vector.
Let $\{\mF_n\}^d = \{\mf = (f_1, \ldots, f_d)^{\top} : f_i \in \mF_n, i = 1, \ldots, d\}$ represent the $d$-fold Cartesian product of $\mF_n$.
As an example, when a $\mathbb{R}^{d}$-valued function $\mf$ belongs to $\{L_2(Q)\}^d$, 
it implies that each $f_i$ is a member of $L_2(Q)$ for all $1 \leq i \leq d$, where $L_2(Q)$ represents the space of all square-integrable functions with respect to the measure $Q$.
$I_A$ represents the indicator function of the set $A$.
For any function \(\mf\) of random elements \(X\), we define \(\dot{\mf}(X) = \mf(X) - \mathbb{E}[\mf(X)]\).
For two functions $\mf_1:\R^p \rightarrow \R^d$ and $\mf_2:\R^d\rightarrow \R^l$,
the composition of $\mf_1$ and $\mf_2$ is represented as $\mf_2\circ \mf_1$.
We denote two sets $\mA_{n, k}$ and $\mC_{n, k}$ as follows:
\begin{equation}\label{PAC} 
    \begin{split}
        \mA_{n, k}&=\left\{\left(\tau_{1}, \ldots, \tau_{k}\right) \in\{1, \ldots, n\}^{k}: \tau_{j} \neq \tau_{l}, \text { for } j \neq l\right\},\\
        \mC_{n, k}&=\left\{\left(\tau_{1}, \ldots, \tau_{k}\right) \in\{1, \ldots, n\}^{k}: \tau_{j} < \tau_{l}, \text { for } j < l\right\}.
    \end{split}
\end{equation}

\subsection{Neural Networks}
Feedforward neural networks (FNNs), commonly referred to as Multi-layer perceptrons, are composed of a series of fully-connected layers. Convolutional neural networks (CNNs) are structured as a sequence that integrates both convolutional and fully-connected layers. These architectures are fundamental and widely utilized in deep learning. 
Given \(W \in \mathbb{R}^{p_2 \times p_1}\) and \(b \in \mathbb{R}^{p_2}\), a fully connected layer \(\FC^{\sigma}_{W,b}: \mathbb{R}^{p_1} \rightarrow \mathbb{R}^{p_2}\) is expressed as \(\FC^{\sigma}_{W,b}(a) = \sigma\left(Wa - b\right)\). 
In this paper, we utilize the ReLU activation function, defined as $\sigma(x) = \max(x, 0)$, which is applied element-wise when $x$ is not a scalar. Additionally, $id(x)=x$ represents the identity function.
For any positive integer $N\in \N_{+}$, we define $[N]\coloneqq \{1,\ldots,N\}$. The function class of FNNs can then be defined as follows:
\begin{definition}[Fully-connected Neural Networks (FNNs)]
        \label{def_FNN}
        Let $\mathcal{L}, \mathcal{N}, \mathcal{S} \in \mathbb{N}_+$ denote the depth, maximum width and total parameters of a FNN. For each layer $i \in [\mL]$, let $k_i$ denote width,
        i.e., weight $W_i\in \R^{k_{i}\times k_{i-1}}$ and intercept vector $b_i\in \R^{k_{i}}$.
        Then \( \mathcal{N} = \max\{k_1, \ldots, k_\mathcal{L}\} \) and \( \mathcal{S} = \sum_{i=0}^{\mathcal{L}} k_{i+1} \times (k_i + 1) \). Let $k_0=p$ and $k_{\mathcal{L}+1}=1$ represent the widths of the input and output layers, respectively.
        We define the class of FNNs by 
        $$\F_{FNN}=\left\{f\mid f=\FC^{id}_{W_\mathcal{L},b_\mathcal{L}}\circ\cdots\circ\FC^\sigma_{W_0,b_0}:\R^p\rightarrow \R\right\}.$$
    \end{definition}

Let $K,H,H^\prime\in \N_{+}$ denote a filter size, input channel size, and output
channel size, respectively.
Then, for a filter $w=$ $\left(w_{n, j, i}\right)_{n \in[K], j \in\left[H^{\prime}\right], i \in[H]}$ $ \in \mathbb{R}^{K \times H^{\prime} \times H}$, 
we define the one-sided padding and stride-one convolution 
as $4$-th order tensor
$L_p^w=\left(\left(L_p^w\right)_{\alpha, i}^{\beta, j}\right) \in \mathbb{R}^{p \times p \times H^{\prime} \times H}$,
$$
\left(L_p^w\right)_{\alpha, i}^{\beta, j}:= \begin{cases}w_{(\alpha-\beta+1), j, i} & \text { if } 0 \leq \alpha-\beta \leq K-1, \\ 0 & \text { otherwise, }\end{cases}
$$
where $i$ (resp. $j$ ) runs through 1 to $H$ (resp. $H^{\prime}$ ) and $\alpha$ and $\beta$ through 1 to $p$. 
Given a fixed input dimension \(p\), we denote the linear mapping as \(L^w\) (omitting subscript \(p\) for simplicity). This mapping operates from \(\mathbb{R}^{p \times H}\) to \(\mathbb{R}^{p \times H^\prime}.\) For \(x = (x_{\alpha,i}) \in \mathbb{R}^{p \times H}\), \(L^w\) transforms \(x\) into \(y = (y_{\beta,j}) \in \mathbb{R}^{p \times H^\prime}\) via:  
$y_{\beta,j} = \sum_{i,\alpha} \left(L^w\right)_{\alpha,i}^{\beta,j} x_{\alpha,i}$.
For a weight tensor \(w \in \mathbb{R}^{K \times H^\prime \times H}\) and a bias vector \(b \in \mathbb{R}^{H\prime}\), the convolutional layer \(\operatorname{Conv}_{w, b}^{\sigma}: \mathbb{R}^{p \times H} \rightarrow \mathbb{R}^{p \times H^\prime}\) is defined as:
$
\operatorname{Conv}_{w, b}^{\sigma}(x) = \sigma\left(L^w(x) - \mathbf{1}_p \otimes b\right),
$
where \(\mathbf{1}_p\) is a \(p\)-dimensional vector of ones and \(\otimes\) denotes the outer product. The ResNet-type CNN is constructed by sequentially concatenating one convolution block, \(\mM\) residual blocks and one fully connected layer.
\begin{definition}[ResNet-type Convolutional Neural Networks(CNNs)]
        \label{def_CNN}
        Let $\mM, \mD, \mH, \mathcal{K}\in \N_{+}$ represent the number of residual 
        blocks and depth, channel size and filter size of blocks, respectively.
        For $m \in[\mM]$ and $l \in[\mL]$, 
        $w_m^{(l)} \in \mathbb{R}^{\mathcal{K} \times \mH \times \mH}$ and $b_m^{(l)} \in \mathbb{R}^\mH$
        denote a weight tensor and bias of the l-th layer of the m-th block in the convolution part, respectively. 
        Additionally, $W \in \mathbb{R}^{1 \times p\mH}$ and $b \in \mathbb{R}$ denote $a$ weight matrix and a bias for the fully-connected layer part, respectively. 
        ResNet-type CNNs is defined as,
        $$
        \begin{aligned}
            \F_{CNN}=\{f\mid f=\mathrm{FC}_{W, b}^{\mathrm{id}} & \circ\left(\operatorname{Conv}_{\boldsymbol{w}_\mM, \boldsymbol{b}_\mM}^\sigma+\mathrm{id}\right) \circ \cdots \\
        & \circ\left(\operatorname{Conv}_{\boldsymbol{w}_1, \boldsymbol{b}_1}^\sigma+\mathrm{id}\right) \circ P:\R^p\rightarrow \R\}
        \end{aligned}
        $$
        Here, $\operatorname{Conv}_{\boldsymbol{w}_m, \boldsymbol{b}_m}^\sigma =  \operatorname{Conv}_{w_m^{(L)}, b_m^{(L)}}^\sigma \circ \cdots \circ \operatorname{Conv}_{w_m^{(1)}, b_m^{(1)}}^\sigma$ and $P$ is a padding operation that aligns the number of channels by adding zeros: $P: \mathbb{R}^p \to \mathbb{R}^{p \times \mH},\; x \mapsto [x\; 0\; \cdots\; 0]$.  
        
    \end{definition}

\section{FCCov: A New Dependence Measure}\label{sec:measure}
   In this section, we introduce Fr\'echet cumulative covariance as a new conditional mean independence measure and corresponding properties and fast computation algorithms.
    
\subsection{Fr\'echet Cumulative Covariance}

    Let \( U \in \mathbb{R} \) and \( V \)  be a random object that takes values in a metric space $(\mV, d)$. Let \( (\bar{U}, \bar{V}) \) and \( (\tilde{U}, \tilde{V}) \) be independent copies of \( (U, V) \), with \( 0 < \operatorname{var}(U) < \infty \) and $E d^2(V,\Tilde V) < \infty$.
    If $E(U \mid V)=E(U)$, then for all $v_0, v_1 \in \operatorname{supp}(V)$, we can derive that
    \begin{equation}\label{equ:equivalent}
        E\left(U \mid d(V,v_0)<d(v_1,v_0)\right)= E\left[E\left(U \mid V\right)\mid d(V,v_0)<d(v_1,v_0)\right] = E(U),
    \end{equation}
    where the equations follow from the smoothness property of conditional expectation and the law of iterated expectations, and  $\operatorname{supp}(V)$ stands for the support of the conditioning variable $V$.
    Inspired by the cumulative covariance (CCov) introduced by \citet{zhou2020model, li2023testing}, we further observe that, for all $v_0, v_1 \in \operatorname{supp}(V)$
    \[
    \begin{aligned}
    E(U | V)=E(U)
    \Longrightarrow & E\left(U \mid d(V,v_0) <d(v_1,v_0)\right)=E(U) \\
    \Longleftrightarrow  &\operatorname{cov}\left\{U, I\left[d(V,v_0)<d(v_1,v_0)\right]\right\}=0 \\
    \Longleftrightarrow  &E\left[\operatorname{cov}^2\left\{U, I[d(V,\Tilde{V})<d(\Bar{V}, \Tilde{V})] \mid \Bar{V}, \Tilde{V}\right\}\right]=0.
    \end{aligned}
    \]
    This motivates us to define the Fr\'echet Cumulative Covariance (FCCov) as follows.
    \begin{definition}\label{def_FCCov}
    Let $(\mV, d)$ be a separable metric space.
    For any random variables $V\in \mV$ and $U\in \R$
    with $\Var(U)<\infty$ and $E d^2(V,\Tilde V) < \infty$.
    The Fr\'echet cumulative covariance \normalfont{FCCov}$(U\mid V)$ is defined by
    \begin{equation}
        \label{def:FCCov}
        \FCCov(U\mid V) = E\left[\operatorname{cov}^2\left\{U, I[d(V,\Tilde{V})<d(\Bar{V},\Tilde{V})] \mid \Bar{V},\Tilde{V}\right\}\right]. 
    \end{equation}
    \end{definition}
    

    The separability assumption is a common hypothesis for metric spaces, ensuring that open balls generate the Borel \(\sigma\)-field of \(\mV\) \citep{van2003probability}.
    One of the most important properties of the CCov$(U\mid V)$ \citep{zhou2020model} is that it is nonnegative and equals zero if and only if $E(U \mid V)=E(U)$. Although the above discussions brings FCCov closer to this property, it remains challenging to deduce $E(U \mid V)=E(U)$ from equation \eqref{equ:equivalent}.
    Let
    \begin{align*}
    S_2 &= \big\{B\left(v_0, d(v_0,v_1)\right): \text{ for all } v_0, 
    v_1 \in \operatorname{supp}(V)\big\}, \\
    \mathscr{A} &= \big\{\text{finite unions and intersections of sets in }S_2\big\}.
    \end{align*}
    We observe that, given the separability of the space, \( S_2 \) generates the Borel \(\sigma\)-field of the metric space. And we define \(\Pi\) which consists of the sets \( A \in \mathscr{A} \) satisfying \( E\left(U \cdot I\{V \in A\}\right) = E(U) \cdot E(I\{V \in A\}) \). We define \(\Lambda\) to consist of the sets \( C \in \sigma(V) \) that satisfy \( E\left(U \cdot I\{V \in C\}\right) = E(U) \cdot E(I\{V \in C\}) \). Then we demonstrate that \(\Pi\) is a \(\pi\)-system and \(\Lambda\) is a \(\lambda\)-system. Utilizing Dynkin's \(\pi\)-\(\lambda\) theorem, we establish the following important theorem for FCCov.
    \begin{theorem}\label{prop_FCCov}
        Let $U$ and $V$ be the random variables defined in Definition \ref{def_FCCov}. 
        Then, \begin{enumerate}
            \item[(a).] $E(U \mid V) \overset{a.s.}{=}E(U) 
            \Longleftrightarrow
            E\left(U \mid d(V,v_0)<d(v_1,v_0)\right)=E(U)$.

            \item[(b).] \normalfont{FCCov}$(U\mid V) \ge 0$. Moreover, $E(U\mid V) \overset{a.s.}{=} E(U) \Longleftrightarrow \text{\normalfont{FCCov}}(U\mid V)=0$. 
        \end{enumerate}
        
    \end{theorem}

    Theorem \ref{prop_FCCov}(b) establishes that \(\FCCov(U \mid V)\) provides a new measure of independence between Euclidean \(U\) and metric space valued \(V\) from a conditional mean perspective.
    Observing the indicator function involved in FCCov, the robustness is another valuable property of FCCov inherited from CCov. This property also enables our nonlinear SDR method introduced in Section \ref{sec:method} to exhibit robustness against outliers.


    For the estimation of  \(\FCCov(U \mid V)\) based on random samples $\{(U_i,V_i),i=1,\ldots n\}$ drawn independently from $(U,V)$, we assume $E[U]=0$ without loss of generality and consider the following estimator.
    \begin{equation}\label{FCCov_sample_form}
        \widehat{\FCCov}(U\mid V) = \frac{1}{(n)_4}\sum_{(i,j,k,l)\in\mathcal{A}_{n,4}}
        U_iU_j\phi(V_i,V_k,V_l)\phi(V_j,V_k,V_l),
    \end{equation}
    where $\phi(V_1, V_2, V_3) = I[d(V_1,V_3)<d(V_2,V_3)]$ and $(n)_m = n(n-1)\cdots (n-m+1), 1 \leq m \leq n$. $\widehat{\FCCov}(U\mid V)$ is an unbiased 4th-order U-statistics. Please refer to Proposition \textcolor{blue}{S.1} in the supplementary material for the detailed derivation. 
    

    \subsection{Fast Computation Algorithm}\label{sec:computation}
    Although we can leverage many desirable statistical properties of U-statistics, the computational complexity of the naive estimator of $\widehat{\FCCov}(U\mid V)$, \(O(n^4)\), makes it impractical for large-scale applications. To address this computational challenge, we propose an efficient algorithm for estimating \({\FCCov}(U \mid V)\) based on the following two preprocessing steps.
    \begin{itemize}
      \item[1] For each \(t = 1, \ldots, n\), we sort all samples based on their distance to \(V_t\), resulting in the ordered sequence, $d(V_{(1)t}, V_t) < d(V_{(2)t}, V_t) < \cdots < d(V_{(n)t}, V_t).$
       \item[2] For each \(V_{(i)t}\), define the centered predictors as \(\dot{U}_{(i)t} = U_{(i)t} - \bar{U}_n\) and \(\dot{U}_t = U_t - \bar{U}_n\). 
    \end{itemize} 
    The following theorem shows how to compute $\widehat{\FCCov}(U \mid V)$ more efficiently.
    \begin{theorem}\label{fast_alg}
      For i.i.d. samples $\{U_i,V_i\}_{i=1}^n$ drawn from the joint distribution of
      $(U,V)$, 
      \begin{equation}
        \begin{split}
        (n)_4\cdot \widehat{\text{\normalfont{FCCov}}}(U \mid V) =& \sum_{t=1}^{n}\sum_{r=1}^{n}\left[\bigl(\sum_{i=1}^{r-1}U_{(i)t}\bigr)^2-\sum_{i=1}^{r-1}U_{(i)t}^2 - 2U_t\cdot \sum_{i=1}^{r-1}U_{(i)t} \right] \\
        &+ 2(n-1)\sum_{i=1}^{n}U_i^2.
        \end{split}
      \end{equation}
    \end{theorem}
    Theorem \ref{fast_alg} guarantees that $\widehat{\FCCov}$ can be computed in only $\mO(n^2 \log n)$ operations.


\section{FCCov-Net for Nonlinear SDR}\label{sec:method}

    \subsection{The FCCov Based Objective Function}
    Let \((\Omega, \mG, P)\) be a probability space and  \((\Omega_X, \mG_X), (\Omega_Y, \mG_Y)\) and $(\Omega_{XY}, \mathcal{G}_{XY})$ be measurable spaces, where \(\Omega_X \subset \mathbb{R}^p\) and \(\Omega_Y \subset \mathcal{Y}\), with \((\mathcal{Y}, d_{\mY})\) being a separable metric space. 
    Assume that \(\Omega_{XY} = \Omega_X \times \Omega_Y\) and \(\mG_{XY} = \mG_X \times \mG_Y\). Let \(X\), \(Y\), and \((X, Y)\) be random elements taking values in \(\Omega_X\), \(\Omega_Y\), and \(\Omega_{XY}\), with distributions \(P_X\), \(P_Y\) and \(P_{XY}\). 
    The \(\sigma\)-field generated by \(X\) is defined as \(\sigma(X) = X^{-1}(\mG_X)\). The conditional distribution of \(X\mid Y\) is denoted by \(P_{X\mid Y}(\cdot \mid \cdot): \mG_X \times \Omega_Y \to \mathbb{R}\).  
    We consider following nonlinear SDR problem
    \begin{equation}\label{conditional_indep}
        Y \indep X \mid \mG_{Y\mid X},
    \end{equation}
    where $\mG_{Y | X}$ is a sub-$\sigma$-field satisfying the conditional independence. \citet{libing_generalSDR} demonstrated that, under mild conditions, the $\sigma$-field $\mG_{Y |X}$ exists and is unique. 
    Assume that  family of probability measures $\{P_{X|Y} (\cdot|y) : y\in \Omega_Y\}$ is dominated by a $\sigma$-finite measure. Following \citet{libing_generalSDR}, 
     there exists a unique minimal sufficient $\sigma$-field (or central $\sigma$-field) $\mG_{Y |X}$ such that \eqref{conditional_indep} holds true. Equivalently speaking, there exists $\mf_0(X)\in\R^d (d<p)$ with $E\big\{{\mf_0}(X)\big\} = 0$ and $\operatorname{Var}\big\{{\mf_0}(X)\big\} = I_d$ satisfying $\sigma\{\mf_0(X)\}=\mG_{Y |X}$. Here, we assume that $d$ is known. However, for practical applications, we propose an empirically effective algorithm to estimate the intrinsic dimension $d$.

    
    Replacing \(\mf_0\) with any one-to-one function transformation does not alter the $\sigma$-field generated by
    \(\mf_0\). Consequently, \(\mf_0\) itself is unidentifiable. 
    Let \(\mathfrak{S}_{\mG_{Y \mid X}}\) be the central class comprising all square-integrable and \(\mG_{Y \mid X}\)-measurable functions, which is unique and identifiable.
    Following \citet{libing_generalSDR, chen2024deep}, we consider the orthogonal direct sum decomposition of $L_2(P_X)=\mathfrak{S}_{\mG_{Y\mid X}} \oplus \mathfrak{S}^\perp_{\mG_{Y\mid X}}$,  where $\mathfrak{S}^\perp_{\mG_{Y\mid X}}$ be the orthogonal complement of $\mathfrak{S}_{\mG_{Y\mid X}}$. 
 For any $f(X)\in L_2(P_X)$ and $\dot f(X)=f(X)-E(f(X))$, Theorem \ref{prop_FCCov} indicates that
    \[
    \FCCov(f(X)\mid Y) = E\left(E\left\{\dot{f}(X)k(I)|
        \Bar{Y},\Tilde{Y}\right\}E\left\{\dot{f}(X)k(I)|
        \Bar{Y},\Tilde{Y}\right\}\right) \geq 0
    \]
    where $k(I)=I\left(d(Y,\Tilde{Y})<d(\Bar{Y},\Tilde{Y})\right)-E\left[I\left(d(Y,\Tilde{Y})<d(\Bar{Y},\Tilde{Y})\right)\mid \Bar{Y},\Tilde{Y}\right]$. 
    The next theorem reveals the fundamental relationship between FCCov and the two function classes $\mathfrak{S}_{\mG_{Y\mid X}}$ and  $\mathfrak{S}^\perp_{\mG_{Y\mid X}}$, providing a novel guiding principle for nonlinear SDR.

\begin{theorem}\label{constrained}
    \begin{itemize}
       \item[(a).]  For any $g \in \mathfrak{S}^\perp_{\mG_{Y\mid X}}$, \normalfont{FCCov}$\big(g(X)\mid Y\big)=0$.
       \item[\emph{(b).}] \emph{Let $m\in \mathbb{N}^+$ be any positive integer. Then the optimal solution $\mf^*=(f_1^*,\ldots, f_m^*)^\top$ of the following objective function}
        \begin{equation}
            \label{prob1}
            \begin{split}
                &\max\limits_{f_t\in \{L_2(P_X)\}} \sum_{t=1}^m {\FCCov}(f_j(X) \mid Y), \\
                &\text{subject to} \quad \Var\{\mf(X)\} = \textbf{I}_m.
            \end{split}
        \end{equation}
        \emph{satisfies $f_j^*\in \mathfrak{S}_{\mG_{Y\mid X}}$, for all $1\leq j\leq m$.} 
    \end{itemize}
\end{theorem}

\begin{remark}
Note that multiplying any solution to \eqref{prob1} by an orthogonal matrix yields another valid solution. Therefore, without loss of generality, we assume that \(\mf^*\) is the solution satisfying 
\[
-E\left[ E\left\{ \dot{f}^*_i(X) k(I) \mid \bar{Y}, \tilde{Y} \right\} E\left\{ \dot{f}^*_j(X) k(I) \mid \bar{Y}, \tilde{Y} \right\} \right] = l_i^*\cdot I(i=j)
\]
Specifically, we reorder the components of \(\mf^*\) using an orthogonal matrix such that for all \(1 \leq i \leq j \leq m\), \(l_i^* \leq l_j^*\). In the context of linear SDR, the solution \(f_1^*\) can be regarded as the first sufficient direction associated with the maximum eigenvalue of certain SDR matrix.

\end{remark}

Theorem \ref{constrained} guarantees that $\sigma(\mf^*)\subseteq \mathcal{G}_{Y|X}$, which is called the unbiasedness in the literature of nonlinear SDR \citep{libing_generalSDR}. This implies that each component of the solution \(\mf^*\) preserves information about \(\mf_0(X)\).
Keep in mind that our primary objective is to identify nonlinear mappings that reduce the dimensionality of $X$ while retaining as much information about $Y$ as possible, rather than precisely recovering $\mf_0$. Compared to the work of \cite{chen2024deep} and \cite{huang2024deep}, our proposal is not only suitable for general metric space valued responses, but is also expected to be more robust against outliers when responses are Euclidean. This is because the objective function of our loss function involves only the computation of the distances $d(Y,\tilde Y)$ and $d(\bar Y, \tilde Y)$, along with the corresponding indicator function. However, the constraint $\Var\{\mf(X)\} = \textbf{I}_m$ of \eqref{prob1} is difficult for implementation, which needs further refinement. 

    \subsection{Tractable Objective Function With Regularization}
    Our Theorem \ref{constrained} holds for any arbitrary $m$. However, noting that $\mf_0$ is an $\R^d$-valued mapping, it is reasonable to choose the solution $\mf^*$ with $m = d$ as an alternative to $\mf_0$.
    To address the intractable variance constraints in \eqref{prob1}, we incorporate the squared Frobenius norm of the difference between \(\Var\{\mf(X)\}\) and the identity matrix \(I_d\), thereby reformulating the problem into an unconstrained Lagrangian objective.
    \begin{equation}\label{unconstrained}
      L_F(\lambda, \mf) = -\sum_{t=1}^d \FCCov(f_t(X)\mid Y) + \lambda \|\text{Var}\{\mf(X)\} - \textbf{I}_d\|_F^2,
    \end{equation}
    where  \(\lambda > 0\) is the regularization parameter. 
    To establish the relationship between the solution of the unconstrained objective function \eqref{unconstrained} and that of the constrained optimization problem \eqref{prob1}, we state the following assumption.
    \begin{assumption}
        \label{ass_eigengap}
        There is a strict gap among between $l_i^*$ and $l_j^*$  for $1 \leq i < j \leq d+1$,
	\[l_i^* = -\text{\normalfont{FCCov}}(f^*_i(X)\mid Y)<-\text{\normalfont{FCCov}}(f^*_j(X)\mid Y) = l_j^*,\] 
    \end{assumption}
    This condition is parallel to commonly used assumption that the nonzero eigenvalues are different in the literature of  linear SDR, which is essential in theoretical analyses involving eigen-decomposition.
    We then present the following theorem, which establishes the equivalence between the solutions of  \eqref{prob1} and \eqref{unconstrained} and also confirms the unbiasedness of soluton to \eqref{unconstrained} for nonlinear SDR.

    \begin{theorem}\label{thm_unconstrained} Let $\lambda>0$ and \(\mf_{\lambda}^*=(f^*_{1,\lambda}, f^*_{2,\lambda}, \ldots, f^*_{d,\lambda})^\top\) be the minimizer of \eqref{unconstrained} and  \(L = \text{diag}(l_1^*, \dots, l_d^*)\)  .  Then
    \begin{itemize}
    \item[(a).] $\mf^*_\lambda = V(\textbf{I}_d-\frac{L}{2\lambda})^{\frac{1}{2}}\mf^*$, where $V$ is an $m\times m$ orthogonal matrix.
    \item[(b).] \(f_{j, \lambda}^* \in \mathfrak{S}_{\sigma \{\mf_0(X)\}}\),  for all \(1 \leq j \leq d\).
    \end{itemize}
    \end{theorem}

 \cite{huang2024deep} verified that the solution to the objective function of distance covariance with a normality constraint is able to recover part of $\mathfrak{S}_{\sigma \{\mf_0(X)\}}$. They further proposed to regularize certain divergences to replace the normality assumption for real implementations. However, the theoretical property of their regularized approach for nonlinear SDR remains unclear. \cite{chen2024deep}  proposed two regularized schemes for their proposed GMDD-Net approach. They only provided a thorough theoretical analysis for the sequential regularization scheme, which is computational intensive compared to our method. Similar to \cite{huang2024deep}, the theoretical properties of the alternative regularization scheme in \cite{chen2024deep} is not fully explored.
    \subsection{Sample Version}
    For the unconstrained objective \eqref{unconstrained}, direct optimization is infeasible due to the unknown distribution of \(Z = (X, Y)\). Therefore, given the dataset \(\mathscr{D}_n = \{ Z_1, Z_2, \dots, Z_n \}\), it is crucial to design an efficient sample estimator to support our optimization algorithm and control the excess risk bound of the solutions.
    We assume that $E \mf(X) = 0$ without loss of generality. Then the unbiased U-statistic form of $L_F(\lambda,\mf)$ can be expressed as follows:
    \begin{equation}\label{4th_sample_form}
        \begin{split}
            L_n(\lambda,\bm{f}) = \frac{1}{(n)_4}\left(\sum_{t=1}^d 
            \widehat{\FCCov}(f_t(X)|Y) + \lambda \sum_{(i,j,k,l)\in
            \mathcal{A}_{n,4}} h_{2,\bm{f}}(Z_{i},Z_{j},Z_{k},Z_{l})
            \right).
        \end{split}
    \end{equation}
    where $h_{2,\bm{f}}$ is the kernel of the unbiased U-statistic for the Lagrange term 
	$\|\text{Var}\{\mf(X)\} - I_d\|_F^2$. And it has the following expression.
	\begin{align*}
	h_{2,\bm{f}}(Z_1,Z_2,Z_3,Z_4)
    &=  \sum_{t=1}^{d}\left[
    \frac{1}{12}\sum_{(i,j)\in \mathcal{A}_{4,2}}f_t^2(X_i)f_t^2(X_j)
    - \frac{1}{2}\sum_{(i)\in\mathcal{A}_{4,1}}f_t^2(X_i)
    \right] + d\\
    &+ \sum_{1\leq s<t\leq d}\frac{1}{6}\sum_{(i,j)\in\mathcal{A}_{4,2}}
    f_s(X_i)f_s(X_j)f_t(X_i)f_t(X_j).
	\end{align*}
    The expression for \(\widehat{\FCCov}\) is provided in \eqref{FCCov_sample_form}. Clearly $L_n(\lambda,\mf)$ is a U-statistic with an implicit kernel. For theoretical convenience, we also derive its explicit kernel representation, which is provided in the supplementary material.
    Furthermore, the fast computational algorithm for \(\widehat{\FCCov}\) can be applied to accelerate the optimization process.
    We further utilize a neural networks function class $\mathcal{F}_n$, such as $\mF_{FNN}$ or $\mF_{CNN}$ as described in Section \ref{sec:preliminary}, to approximate smooth functions in $L_2(P_X)$. And our final FCCov-Net estimator for nonlinear SDR is defined as
    \begin{equation}\label{equ:sample}
        \hat{\mf}_\lambda=\argmin\limits_{\mf\in\{\mathcal{F}_n\}^d}  L_n(\lambda, \mf).
    \end{equation}

    \section{Nonasymptotic Convergence Rate}\label{sec:theory}
    This section investigates the non-asymptotic properties of our proposed nonlinear SDR method.
    We begin by considering the excess risk defines as
    \begin{equation*}
        R(\lambda, \Hat{\mf}_\lambda) = L_F(\lambda, \Hat{\mf}_\lambda)
        - L_F(\lambda, \mf_\lambda^*),
    \end{equation*}
    where $\mf_\lambda^* = (f_{1,\lambda}^*,...,f_{d,\lambda}^*)$ and $\Hat{\mf}_\lambda = (\Hat{f}_{1,\lambda},...,\Hat{f}_{d,\lambda})$ denote the minimizers of the objective function \eqref{unconstrained} and sample-based problem \eqref{equ:sample}, respectively.
    The excess risk quantifies the discrepancy between the estimated directions and their corresponding ideal counterparts, reflecting the accuracy of the estimated directions. Keep in mind that multiplying $\mf_\lambda^*$ by an $d\times d$ orthonormal matrix from the left will also satisfy the unbiased estimation property, we also define the following risk for theoretical investigation,
      \begin{equation*}
 \kappa^2(\Hat{\mf}_\lambda,{\mf}^*_\lambda) = \min\limits_{Q\in O(d)}\Vert \Hat{\mf}_\lambda -Q {\mf}^*_\lambda
    \Vert^2,
    \end{equation*}
where $O(d)$ is the class consisting of all orthonormal matrices in $\mathbb{R}^{d\times d}$. Before presenting the non-asymptotic results, we introduce some necessary assumptions.
    \begin{assumption}
        \label{ass:bounded}
		There exists an absolute constant $B>1$ such that $\|{f}^{*}_{j, \lambda}\|_{\infty} \leq B$ and $\|f\|_{\infty} \leq B$ for any $1 \leq j \leq d$ and $f \in \mF_n$. 
    \end{assumption}
    
    \begin{assumption}
        \label{ass:holder} ${f}^{*}_{j, \lambda}$ is $\beta$-H\"older function for all $1 \leq j \leq m$, where a $\beta$-H\"older function $f$ is defined as  
        \[
        \|f\|_\beta := \sum_{0 \leq |\alpha| < \lfloor \beta \rfloor} \|\partial^\alpha f\|_\infty + \sum_{|\alpha| = \lfloor \beta \rfloor} \sup_{x \neq y} \frac{|\partial^\alpha f(x) - \partial^\alpha f(y)|}{|x - y|^{\beta - \lfloor \beta \rfloor}} < \infty,
        \]
        Here, $\alpha=\left(\alpha_1, \ldots, \alpha_D\right)$ represents a multi-index, and $\lfloor \beta \rfloor$ denotes the greatest integer less than or equal to $\beta$. Specifically, $\partial^\alpha f:=$ $\frac{\partial^{|\alpha|} f}{\partial x_1^{\alpha_1} \ldots \partial x_D^{\alpha_D}}$, where $|\alpha|:=\sum_{i=1}^D \alpha_i$.
    \end{assumption}

    \begin{assumption}\label{ass:neural network} The neural networks functional classes $\mF_{CNN}$ and $\mF_{FNN}$ satisfy
    \begin{itemize}
    \item[$\diamond$] $\mF_{CNN}$ with depth \(\mD = \mathcal{O}(\log \mM)\), number of channels \(\mH = \mathcal{O}(1)\), and convolutional kernel size \(\mathcal{K} \in \{2, \ldots, p\}\). We choose $\mM = n^{\frac{p}{2\beta + p}}$ and $n>\mM \log(\mM)$.
    \item[$\diamond$] $\mF_{FNN}$ with width $\mathcal{N} = 3^{p+3} \max\big(p \lfloor N^{1/p} \rfloor, N+1\big)$ where $N\in\mathbb{N}_+$ is any positive integer. We choose depth $\mathcal{L} = 12n^{\frac{p}{2(p + 2\beta)}} + 14+ 2p$ and $n>\mL \log(\mL)$.
    \end{itemize}
    \end{assumption}
    Assumptions \ref{ass:bounded} and \ref{ass:holder} are standard conditions frequently employed in the theoretical analysis of neural network error bounds, as demonstrated in \citet{Schmidt_Hieber_2020}. For the $\beta$-H\"older class, the parameter $\beta$ characterizes the smoothness of the function, with larger values of $\beta$ indicating smoother functions. To effectively approximate $\beta$-H\"older continuous functions while balancing the statistical error, we introduce Assumption \ref{ass:neural network}. This condition is also adopted in \citet{shen_app, nearly_tight} and \citet{oono2019approximation}, for studying the approximation capabilities and complexities of FNNs and CNNs. Moreover,  $\mL \log(\mL)$ and $\mM \log(\mM)$ are actually the VC dimensions of FNNs and CNNs respectively.

    By applying the Hoeffding decomposition of U-statistics, we can divide $R(\lambda, \Hat{\mf}_\lambda) $ into several terms.
    The leading term is a sum of independent terms, which can be rigorously analyzed using classical empirical process theory. Additionally, we can bound the variance of this term using the previously defined orthogonal invariant distance $\kappa$.
    Under the aforementioned assumptions, the remainders of the Hoeffding decomposition become negligible compared to the first term, as ensured by Theorem 8.3 in \citet{massart2007concentration}.
    The following lemma establishes the upper bound of $R(\lambda, \Hat{\mf}_\lambda)$.
    \begin{lemma}
        \label{fast_rate}
        Under assumptions\ref{ass_eigengap}-\ref{ass:neural network}, let $\lambda>0$, then for any $\delta>0$, the following bounds hold with probability at least $1 - \delta$.
        \begin{itemize}
            \item[(a).] For $\mF_n=\mF_{FNN}$,
            \[R(\lambda, \Hat{\mf}_\lambda)=
            \mathcal{O} \left( \big(\mathcal{N}\mathcal{L}\big)^{-\frac{4 \beta}{p}}+\frac{\mathcal{S} \mathcal{L}\log\left(\mathcal{S}\right)}{n}\log\frac{n}{\mathcal{S} \mathcal{L}\log\left(\mathcal{S}\right)}  + \frac{\mathcal{S} \mathcal{L}\log\left(\mathcal{S}\right) \log(1/\delta)}{n} \right).
            \]
            \item[(b).] For $\mF_n=\mF_{CNN}$,
            \[R(\lambda, \Hat{\mf}_\lambda)=
            \mO\left(\mM^{-\frac{2\beta}{p}} + \frac{\mM\log \mM}{n}\log \frac{n}{\mM\log \mM} + \frac{\mM\log \mM}{n}\log(1/\delta)\right).
            \]
        \end{itemize}
        
    \end{lemma}

    Finally, by integrating Lemma \ref{fast_rate} and selecting the sample size $n$ to balance the approximation error and the statistical error, we achieve the nonasymptotic error rate for our proposed FCCov-Net, as stated in Theorem \ref{cor_NN}.

\begin{theorem}\label{cor_NN} Under assumptions \ref{ass_eigengap}-\ref{ass:neural network}, let $\lambda>0$, then for any \(\delta > 0\), we have
    \begin{itemize}
    \item[(a).]
      $R(\lambda, \Hat{\mf}_\lambda) = \mathcal{O}\left(n^{-\frac{2 \beta}{p + 2 \beta}} \{1 + \log(1/\delta)\} \log^2 n\right)$ holds true with probability greater than \(1 - \delta\).
     \item [(b).]    $E \kappa^2\left(\hat{\bm{f}_\lambda}, \bm{f}_{\lambda}^*\right)=
      \mathcal{O}\left(n^{-\frac{2 \beta}{p+2 \beta}} \log^2 n\right).$
      \end{itemize}
\end{theorem}

    
    
    Compared to existing theoretical results of nonlinear Fr\'echet SDR methods, such as that of \cite{ying2022frechet}, our theory offers several advantages. First, we derive a non-asymptotic excess risk convergence rate, while \cite{ying2022frechet} only provided the asymptotic results.
    We also obtain a probabilistic upper bound on the distance between the estimated and true directions.
    This provides theoretical insights into the sample size required to achieve the desired accuracy in practice.
    Second, our method achieves a convergence rate of $n^{-\frac{2\beta}{p + 2\beta}}\log^2 n$, which is comparable to the minimax rate for estimating H\"older continuous functions in nonparametric regression with Euclidean response. 
    Compared to the theoretical results of \cite{chen2024deep}, which are limited to Euclidean responses and FNNs, our theoretical framework extends to both Euclidean and non-Euclidean responses, as well as to both FNNs and CNNs, demonstrating significantly broader applicability.

    \section{Simulation}\label{sec:simulation}
    In this section, we evaluate the performance of our proposed FCCov-Net through comprehensive simulation studies where the responses are either in Euclidean space or complex random objects such as probability distributions, symmetric positive definite matrices, and spherical data. For comparison in the Euclidean scenario, we include the Generalized Martingale Difference Divergence based on the Frobenius norm (GMDDNet-F) \citep{chen2024deep}, Deep Dimension Reduction (DDR) \citep{huang2024deep}, and the Generalized Sliced Inverse Regression (GSIR) \citep{libing_generalSDR}.
    We exclude the successive optimization method proposed in \citet{chen2024deep} owing to its marginal performance difference relative to GMDDNet-F and its computational inefficiency.
    GMDDNet-F and DDR are neural network based methods. GSIR, on the other hand, is a nonlinear SDR method based on reproducing-kernel Hilbert Space (RKHS). 
    Additionally, we introduce outliers into some Euclidean data to examine robustness of each method. For scenarios involving complex random objects, we compare our method with the nonlinear Weighted Inverse Regression Ensemble (WIRE) method introduced by \citet{ying2022frechet}, which is also a kernel-based approach.
    To evaluate the computational efficiency of the neural network approach, we select several scenarios and compared the time consumption of our method compared with GSIR and WIRE across varying sample sizes. The results indicate that the neural network approach FCCov-Net significantly outperforms traditional kernel methods in terms of computational speed, substantially reducing the computation time. For detailed results, please refer to the supplementary material.

    To effectively train neural networks in PyTorch, we set the hyperparameters as follows: the batch size is 100, and each epoch consists of 100 iterations. We use "Adam" with default parameters for optimization. The learning rate is set as $0.001$.
    Our method employs a simple fully connected network architecture with layer widths of $p,2^{\lfloor\log p\rfloor+1},2^{\lfloor\log p\rfloor+2},2^{\lfloor\log p\rfloor+1},$ $\ldots,2^4,d$, where $p$ and $d$ represent input and output dimension respectively.
    Unless otherwise specified, all training procedures are conducted based on these settings.

    \subsection{Euclidean Responses} 
    We consider the following two models:
    \begin{flalign*}
        &\textbf{Model I:} \quad Y = \left(\frac{x_1^2}{1+(0.1+0.5x_2)^2}, 0, 0\right)^\top
        + \bm{\varepsilon}_1, & \\
        &\textbf{Model II:} \quad Y = \left(\frac{x_3}{x_4+2}, x_1^2\right)^\top + \bm{\varepsilon}_2, &
    \end{flalign*}
    where $\bm{\varepsilon}_1 \sim 0.25\cdot N(\bm{0}, \textbf{I}_3)$ and $\bm{\varepsilon}_2\sim 0.1\cdot N(\bm{0}, \textbf{I}_2)$
    are independent of $X$.
    Additionally, we consider three different distributional scenarios for the $p$-dimensional predictor vector $X = (x_1, \ldots, x_p)^\top$:
    \begin{flalign*}
        &\textbf{(A):} \quad  X \sim N(\bm{0}, \textbf{I}_p), &\\
        &\textbf{(B):} \quad  X \sim N(\bm{0}, \Sigma), \text{where $\Sigma=(0.5^{|i-j|})$ 
        for $i,j=1,\ldots,p$}, &\\
        &\textbf{(C):} \quad  X \sim U([-2,2]^p), \text{where $[-2,2]^p=\underbrace{[-2,2]\times[-2,2]\times\cdots\times[-2,2]}_{p}$ }, &
    \end{flalign*}
    where $\textbf{I}_p$ is the identity matrix and $U([-2,2]^p)$ is multivariate uniform distribution.
    The true sufficient predictors for models II and III are 
    \[
        \frac{x_1^2}{1+(0.1+0.5x_2)^2} \text{\hspace{1cm} and \hspace{1cm}} \begin{pmatrix}
            \frac{x_3}{x_4+2} \\
            x_1^2
        \end{pmatrix}
    \] respectively. Thus, the structural dimensions of the nonlinear SDR are $d = 1$ for Model I and $d = 2$ for Model II.
    

    For each combination of settings and distributional scenarios, we conducted experiments under $p = 10, 20$ and $n = 500, 800, 1000$. To assess the performance of these methods, we utilize the distance correlation $\rho\{f(X), \hat{f}(X)\}$ proposed by \citet{szekely2007measuring} between the true sufficient predictors and their estimators. A larger value of $\rho$ indicates better estimation. To compute the distance correlations, an independent testing sample comprising 20\% of the training sample size is generated. To mitigate the effects of randomness, this procedure is repeated 100 times. The mean and standard deviation of the distance correlations from these 100 repetitions are computed as the evaluation criteria.
    
    \begin{table}[htbp]
        \centering
        \small
        \renewcommand{\arraystretch}{0.58} 
        \caption{Mean (standard deviation) of distance correlations with true predictor based on 100 repetitions.}
        \label{tab:Euclidean}
        \scalebox{0.9}{
        \begin{tabular}{@{\hspace{10pt}}c@{\hspace{10pt}}c@{\hspace{10pt}}c@{\hspace{5pt}}c@{\hspace{15pt}}c@{\hspace{15pt}}c@{\hspace{15pt}}c@{\hspace{15pt}}c@{\hspace{15pt}}}
            \toprule
            \multicolumn{3}{c}{Models} & & \multicolumn{4}{c}{Distance correlation with true predictor} \\ 
            \cmidrule(r){1-3} \cmidrule(r){5-8} 
            $X$ & $Y|X$ & $(n,p)$ & & FCCov-Net & GMDDNet-F & GSIR & DDR \\  
            \midrule
            A  & I & (500, 10) & & \textbf{0.930(0.016)}  & 0.905(0.102)  & 0.929(0.021)  & 0.878(0.159)\\
                & & (800, 10) & & \textbf{0.941(0.009)}  & 0.938(0.010)  & 0.947(0.016)  & 0.924(0.129)\\
                & & (1000, 10) & & 0.945(0.009)  & 0.944(0.008)  & 0.952(0.011)  & \textbf{0.957(0.036)}\\
                & & (500, 20) & & \textbf{0.807(0.035)}  & 0.701(0.177)  & 0.712(0.074)  & 0.792(0.208)\\
                & & (800, 20) & & 0.841(0.025)  & 0.850(0.072)  & 0.787(0.052)  & \textbf{0.904(0.112)}\\
                & & (1000, 20) & & 0.855(0.021)  & 0.880(0.040)  & 0.815(0.041)  & \textbf{0.929(0.088)}\\
            \noalign{\vskip 5pt}
           & II & (500, 10) & & \textbf{0.775(0.131)}  & 0.765(0.186)  & 0.743(0.130)  & 0.694(0.190)\\
            & & (800, 10) & & \textbf{0.763(0.134)}  & 0.758(0.178)  & 0.731(0.129)  & 0.751(0.157)\\
            & & (1000, 10) & & 0.763(0.153)  & 0.741(0.208)  & 0.732(0.149)  & \textbf{0.788(0.171)}\\
            & & (500, 20) & & \textbf{0.666(0.123)}  & 0.690(0.185)  & 0.589(0.106)  & 0.562(0.182)\\
            & & (800, 20) & & \textbf{0.719(0.102)}  & 0.729(0.191)  & 0.631(0.091)  & 0.699(0.172)\\
            & & (1000, 20) & & 0.698(0.151)  & 0.689(0.224)  & 0.611(0.134)  & \textbf{0.710(0.195)}\\
        \noalign{\vskip 5pt}
            \midrule
        B  & I & (500, 10) & & \textbf{0.937(0.014)}  & 0.916(0.019)  & 0.825(0.043)  & 0.864(0.179)\\
            & & (800, 10) & & \textbf{0.947(0.009)}  & 0.943(0.009)  & 0.859(0.027)  & 0.922(0.151)\\
            & & (1000, 10) & & 0.950(0.010)  & 0.948(0.009)  & 0.862(0.028)  & \textbf{0.951(0.077)}\\
            & & (500, 20) & & \textbf{0.823(0.036)}  & 0.749(0.159)  & 0.577(0.080)  & 0.791(0.224)\\
            & & (800, 20) & & 0.854(0.022)  & 0.862(0.058)  & 0.636(0.067)  & \textbf{0.922(0.083)}\\
            & & (1000, 20) & & 0.869(0.019)  & 0.895(0.030)  & 0.670(0.052)  & \textbf{0.913(0.142)}\\
        \noalign{\vskip 5pt}
        & II & (500, 10) & & \textbf{0.774(0.118)}  & 0.708(0.190)  & 0.713(0.103)  & 0.696(0.147)\\
        & & (800, 10) & & \textbf{0.716(0.198)}  & 0.638(0.243)  & 0.658(0.178)  & 0.664(0.211)\\
        & & (1000, 10) & & \textbf{0.717(0.186)}  & 0.685(0.216)  & 0.653(0.164)  & 0.695(0.191)\\
        & & (500, 20) & & \textbf{0.703(0.105)}  & 0.678(0.162)  & 0.632(0.087)  & 0.539(0.154)\\
        & & (800, 20) & & \textbf{0.721(0.118)}  & 0.679(0.178)  & 0.630(0.099)  & 0.621(0.167)\\
        & & (1000, 20) & & \textbf{0.714(0.135)}  & 0.664(0.186)  & 0.621(0.110)  & 0.677(0.171)\\
        \noalign{\vskip 5pt}
        \midrule
        C  & I & (500, 10) & & \textbf{0.951(0.009)}  & 0.876(0.155)  & 0.855(0.035)  & 0.877(0.179)\\
            & & (800, 10) & & \textbf{0.968(0.006)}  & 0.920(0.129)  & 0.897(0.017)  & 0.934(0.122)\\
            & & (1000, 10) &  & \textbf{0.973(0.004)}  & 0.956(0.006)  & 0.911(0.015)  & 0.969(0.009)\\
            & & (500, 20) & & \textbf{0.831(0.040)}  & 0.595(0.159)  & 0.477(0.084)  & 0.797(0.177)\\
            & & (800, 20) & & 0.885(0.021)  & 0.799(0.132)  & 0.543(0.065)  & \textbf{0.910(0.063)}\\
            & & (1000, 20) & & 0.900(0.017)  & 0.862(0.085)  & 0.604(0.052)  & \textbf{0.935(0.077)}\\
        \noalign{\vskip 5pt}
        & II & (500, 10) & & \textbf{0.663(0.149)}  & 0.553(0.171)  & 0.625(0.123)  & 0.566(0.114)\\
        & & (800, 10) & & \textbf{0.661(0.162)}  & 0.507(0.172)  & 0.623(0.153)  & 0.561(0.131)\\
        & & (1000, 10) & & \textbf{0.667(0.154)}  & 0.538(0.174)  & 0.616(0.136)  & 0.592(0.117)\\
        & & (500, 20) & & 0.525(0.110)  & 0.496(0.150)  & \textbf{0.551(0.100)}  & 0.465(0.107)\\
        & & (800, 20) & & \textbf{0.593(0.134)}  & 0.489(0.158)  & 0.557(0.113)  & 0.502(0.133)\\
        & & (1000, 20) & & \textbf{0.590(0.141)}  & 0.489(0.161)  & 0.536(0.119)  & 0.517(0.130)\\
    \noalign{\vskip 5pt}
        \bottomrule
        \end{tabular}
        }
    \end{table}

    From Table \ref{tab:Euclidean}, our proposed method consistently outperforms 
    the other three approaches in the majority of cases.
    Next, to illustrate the robustness of our method in the presence of outliers, we follow \citet{zhang2019robust, zhang2021robust} to modify the distributions of 
    $X$ and the error $\bm{\varepsilon}$ term , which are stated as follows:

    \textbf{Outlier case-1}
    We generate \( n \) samples of \( X \sim N(0, \Sigma) \), where \( \Sigma = (\sigma_{ij}) \) with \( \sigma_{ij} = 0.5^{|i - j|} \) for \( i, j = 1, 2, \ldots, p \). Then, we randomly replace \( r\% \) of the observations of \( x_1 \) with \( 2 \cdot t(1) \), where \( t(1) \) denotes the t-distribution with one degree of freedom. 

    \textbf{Outlier case-2}
    Let $\bm{\varepsilon}_i \sim 0.25\cdot N(\bm{0},\textbf{I})$, $i=1,2$ and $X$ satisfies \textbf{(B)}.
    Subsequently, we introduce outliers 
    by randomly replacing the distribution of the error 
    term \( \varepsilon \) with \( r\% \) of values drawn from a uniform 
    distribution \( U(-50, 50) \).

    Where $r\%$ is set to be $10\%$, $30\%$ and $50\%$ and the the means and standard errors of 
    distance correlations of three methods are given in Table \ref{tab:outliers_1} and \ref{tab:outliers_2},
    \begin{table}[htbp]
        \centering
        \small
        \renewcommand{\arraystretch}{0.6} 
        \caption{Distance correlation for \textbf{outlier case-1} when predictors have outliers.}
        \label{tab:outliers_1}
        \scalebox{0.9}{
        \begin{tabular}{@{\hspace{8pt}}c@{\hspace{8pt}}c@{\hspace{8pt}}c@{\hspace{5pt}}c@{\hspace{12pt}}c@{\hspace{12pt}}c@{\hspace{12pt}}c@{\hspace{12pt}}c@{\hspace{12pt}}}
            \toprule
            \multicolumn{3}{c}{Models} & & \multicolumn{4}{c}{Distance correlation with true predictor} \\ 
            \cmidrule(r){1-3} \cmidrule(r){5-8} 
            outlier(r\%) & $Y|X$ & $(n,p)$ & & FCCov-Net & GMDDNet-F & GSIR & DDR \\ 
            \midrule
            10\% & I & (1000, 10) & & \textbf{0.940(0.097)}  & 0.297(0.098)  & 0.553(0.224)  & 0.633(0.202)\\ 
            & I & (3000, 20) & & \textbf{0.883(0.163)}  & 0.148(0.062)  & 0.351(0.192)  & 0.691(0.188)\\
            & II & (1000, 10) & & \textbf{0.667(0.185)}  & 0.329(0.086)  & 0.585(0.199)  & 0.591(0.146)\\
            & II & (1000, 20) & & \textbf{0.662(0.142)}  & 0.301(0.091)  & 0.521(0.191)  & 0.526(0.153)\\
            \noalign{\vskip 5pt} 
            30\% & I & (1000, 10) & & \textbf{0.905(0.180)}  & 0.171(0.071)  & 0.240(0.155)  & 0.254(0.106)\\
              & I & (3000, 20) & & \textbf{0.906(0.148)}  & 0.111(0.051)  & 0.134(0.083)  & 0.255(0.124)\\
              & II & (1000, 10) & & \textbf{0.632(0.228)}  & 0.293(0.090)  & 0.381(0.155)  & 0.326(0.101)\\
              & II & (1000, 20) & & \textbf{0.548(0.166)}  & 0.202(0.045)  & 0.339(0.142)  & 0.276(0.081)\\
            \noalign{\vskip 5pt} 
            50\% & I & (1000, 10) & & \textbf{0.946(0.083)}  & 0.162(0.059)  & 0.170(0.067)  & 0.172(0.065)\\
            & I & (1000, 20) & & \textbf{0.824(0.145)}  & 0.132(0.038)  & 0.150(0.061)  & 0.147(0.044)\\
            & II & (1000, 20) & & \textbf{0.573(0.203)}  & 0.208(0.051)  & 0.265(0.087)  & 0.231(0.053)\\
            & II & (3000, 20) & & \textbf{0.524(0.264)}  & 0.183(0.056)  & 0.184(0.049)  & 0.195(0.070)\\
            \bottomrule
          \end{tabular}
        }
    \end{table}
    
    \begin{table}[htbp]
        \centering
        \small
        \renewcommand{\arraystretch}{0.6} 
        \caption{Distance correlation for \textbf{outlier case-2} when predictors have outliers.}
        \label{tab:outliers_2}
        \scalebox{0.9}{
        \begin{tabular}{@{\hspace{8pt}}c@{\hspace{8pt}}c@{\hspace{8pt}}c@{\hspace{5pt}}c@{\hspace{12pt}}c@{\hspace{12pt}}c@{\hspace{12pt}}c@{\hspace{12pt}}c@{\hspace{12pt}}}
            \toprule
            \multicolumn{3}{c}{Models} & & \multicolumn{4}{c}{Distance correlation with true predictor} \\ 
            \cmidrule(r){1-3} \cmidrule(r){5-8} 
            outlier(r\%) & $Y|X$ & $(n,p)$ & & FCCov-Net & GMDDNet-F & GSIR & DDR \\ 
            \midrule
            10\% & I & (1000, 10) & & \textbf{0.930(0.010)}  & 0.512(0.270)  & 0.851(0.025)  & 0.254(0.106)\\ 
            & I & (3000, 20) & & \textbf{0.812(0.022)}  & 0.252(0.096)  & 0.723(0.026)  & 0.224(0.094)\\
            & II & (1000, 10) & & \textbf{0.702(0.160)}  & 0.586(0.184)  & 0.662(0.151)  & 0.343(0.089)\\
            & II & (3000, 10) & & \textbf{0.734(0.138)}  & 0.634(0.197)  & 0.682(0.131)  & 0.371(0.114)\\
            \noalign{\vskip 5pt} 
            30\% & I & (1000, 10) & & \textbf{0.887(0.019)}  & 0.230(0.096)  & 0.701(0.069)  & 0.206(0.080)\\
              & I & (3000, 10) & & \textbf{0.918(0.010)}  & 0.384(0.136)  & 0.817(0.028)  & 0.247(0.122)\\
              & II & (1000, 10) & & \textbf{0.568(0.128)}  & 0.346(0.076)  & 0.498(0.101)  & 0.322(0.083)\\
              & II & (3000, 10) & & \textbf{0.636(0.161)}  & 0.397(0.149)  & 0.581(0.146)  & 0.308(0.090)\\
            \noalign{\vskip 5pt} 
            50\% & I & (1000, 10) & & \textbf{0.700(0.105)}  & 0.184(0.077)  & 0.272(0.123)  & 0.201(0.078)\\
            & I & (3000, 10) & & \textbf{0.852(0.019)}  & 0.196(0.079)  & 0.566(0.101)  & 0.187(0.109)\\
            & II & (1000, 10) &  & 0.320(0.075)  & 0.259(0.062)  & \textbf{0.332(0.076)}  & 0.305(0.080)\\
            & II & (3000, 10) & & \textbf{0.399(0.120)}  & 0.290(0.097)  & 0.390(0.101)  & 0.290(0.093)\\
            \bottomrule
          \end{tabular}
        }
    \end{table}

    GMDDNet-F and DDR nearly entirely lose their ability to identify the nonlinear SDR directions when outliers are present in $X$.
    In contrast, the GSIR method demonstrates relatively strong performance across certain models, though it still suffers some reduction in statistical power. Our method, however, remains relatively stable with outliers, exhibiting high robustness.
    We also assess the effectiveness of our proposed dimensionality determination algorithm, which accurately identifies the structural dimension in nine out of ten trials. Further details are provided in the supplementary material.

    \subsection{Complex Metric Space Valued Responses}
    In this section, we compare our method with nonlinear WIRE across two Fr\'echet regression scenarios. 
    We generate \( n \) i.i.d. samples \( \{(X_k, Y_k)\}_{k=1}^n \), where \( X_k \) is a \( p \)-dimensional random vector with independent components uniformly distributed on \( [0, 1] \). The responses \( Y_k \) are generated according to following two scenarios.

    \subsubsection{Distributions responses}
    Let $(\Omega, d_w)$ be the metric space of probability distributions on $\R$ with 
    finite second order moments, equipped with the quadratic Wasserstein metric $d_w$. 
    For distributions $Y_1, Y_2 \in \Omega$, the squared Wasserstein distance is defined as
    \begin{equation}
        \label{Wasserstein}
        d_w^2(Y_1,Y_2) = \int_{0}^{1} \left(Y_1^{-1}(t)-Y_2^{-1}(t)\right) dt
    \end{equation}
    where $Y_1^{-1}$ and $Y_2^{-1}$ are their respective quantile functions.

    We generate random normal distribution $Y$ with quantile function
    $Q_Y(t) = \mu_Y + \sigma_Y\Phi(t)$, where $\Phi(\cdot)$ is the standard normal cumulative distribution function.
    Let $\beta_1 = (0.75,0.25,0,...,0)^\top$, $\beta_2=(0.25, 0.75, 0,...,0)^\top$.
    
    \textbf{Setting I-1}: 
    $\mu_Y\mid X \sim N\{D(X), 0.1^2\}$ and $\sigma_Y = 1$,
    where $D(X) = \sin\Big(4\pi\beta_1^\top X \cdot (2\beta_2^\top X-1)\Big)$.
    \textbf{Setting I-2}: 
    $\mu_Y\mid X \sim N\{D_1(X), 0.1^2\}$ and $\sigma_Y = |D_2(X)| $,
    where $D_1(X) = \left(x_1^2+x_2^2\right)^{\frac{1}{2}} \log\left(x_1^2+x_2^2\right)^{\frac{1}{2}}$,
    $D_2(X) = \sin\left(0.1\pi(x_4+x_5)\right) + x_4^2$.
    
    In Setting I-1, only the mean $\mu_X$ is related to $X$, whereas in Setting I-2, both the mean $\mu_X$ and standard deviation $\sigma_X$ depend on $X$. These settings correspond to structural dimensions $d=1$ and $d=2$, respectively. For the computation of Wasserstein distance \eqref{Wasserstein}, each quantile function of $Y_i$ is discretized into $21$ equispaced points on $[0, 1]$. 
    The simulation results in Table \ref{distribution} demonstrate that our method outperforms the WIRE.

    \begin{table}[htbp]
        \centering
        \renewcommand{\arraystretch}{0.6} 
        \caption{Distance correlations for models with distributional responses.}
        \label{distribution}
        \scalebox{0.9}{
        \begin{tabular}{@{\hspace{25pt}}c@{\hspace{25pt}}c@{\hspace{10pt}}c@{\hspace{25pt}}c@{\hspace{25pt}}c@{\hspace{25pt}}}
            \toprule
            Models & $(p,n)$ &  & FCCov-Net & WIRE \\
            \midrule
            & (10, 1000) & & \textbf{0.723(0.010)} & 0.450(0.002)   \\
            & (10, 2000) & & \textbf{0.922(0.001)} & 0.466(0.001) \\
            \textbf{Setting I-1}& (10, 5000) &  & \textbf{0.976(0.001)} & 0.467(0.000) \\
            & (30, 1000) &  & 0.296(0.005) & \textbf{0.394(0.005)} \\
            & (30, 2000) &  & \textbf{0.524(0.022)} & 0.405(0.001) \\
            & (30, 5000) &  & \textbf{0.841(0.079)} & 0.433(0.000) \\
            \\
            & (10, 1000) & & \textbf{0.937(0.000)} & 0.928(0.000)   \\
            & (10, 2000) & & \textbf{0.938(0.000)} & 0.927(0.000)   \\
            \textbf{Setting I-2}& (10, 5000) &  & \textbf{0.935(0.000)} & 0.926(0.000) \\
            & (30, 1000) &  & \textbf{0.933(0.000)} & 0.922(0.000) \\
            & (30, 2000) &  & \textbf{0.935(0.000)} & 0.923(0.000) \\
            & (30, 5000) &  & \textbf{0.937(0.000)} & 0.920(0.000) \\
            \bottomrule
          \end{tabular}
        }
    \end{table}

    \subsubsection{Symmetric positive-definite matrices responses}
    Let $(\Omega, d_\omega)$ be the metric space $\mS_m^+$ of $m\times m$ symmetric positive-definite matrices endowed with metric $d_\omega$. 
    Although many metrics exist, this section primarily focuses on the Log-Cholesky metric \citep{lin2019riemannian} and the affine-invariant metric \citep{moakher2005differential}.

    To define the aforementioned metrics, we first introduce the exponential and logarithmic operations for matrices.
    For an \( m \times m \) symmetric matrix \( A \), the matrix exponential is defined as \( \exp(A) = I_m + \sum_{j=1}^{\infty} \frac{1}{j!} A^j \), which yields a symmetric positive-definite matrix. Conversely, for a symmetric positive-definite matrix \( Y \), the matrix logarithm is defined as \( \log(Y) = A \), such that \( \exp(A) = Y \). Therefore, for two symmetric positive-definite matrices \( Y_1 \) and \( Y_2 \), the affine-invariant distance can be formulated as follows: 
    \begin{equation} \label{Affine} 
    d_A(Y_1, Y_2) = \left\Vert \log\left(Y_1^{-1/2} Y_2 Y_1^{-1/2}\right) \right\Vert_F \end{equation}

    Similarly, for $Y_1$ and $Y_2$, we denote their Cholesky decompositions as \( Y_i = P_i P_i^\top \) for \( i = 1, 2 \).
    Let \( \lfloor Y \rfloor \) denote the strictly lower triangular part of \( Y \), and let \( \mathbb{D}(Y) \) denote the diagonal part of \( Y \).
    Then, we define the Log-Cholesky distance as
    \begin{equation}
    \label{Log-Cholesky}
    d_L(Y_1, Y_2) = \left\{ \left\| \lfloor P_1 \rfloor - \lfloor P_2 \rfloor \right\|_F^2 + \left\| \log \mathbb{D}(P_1) - \log \mathbb{D}(P_2) \right\|_F^2 \right\}^{1/2}
    \end{equation}

    We generate \( Y \) from a symmetric matrix-variate normal distribution \citep{zhang2021dimension}. In the simplest case, an \( m \times m \) symmetric matrix \( A \) follows a matrix-variate normal distribution \( A \sim N_{mm}(M, \sigma^2) \) if \( A = \sigma Z + M \), where \( M \) is an \( m \times m \) symmetric matrix and \( Z \) is an \( m \times m \) symmetric random matrix with independent \( N(0, 1) \) diagonal elements and \( N(0, 1/2) \) off-diagonal elements. Let \( \beta = (0.5, 0.5, 0, \ldots, 0)^\top \). We consider the following settings where \( Y \) is a symmetric positive-definite matrix.

    \textbf{Setting II-1}:
    \[\log(Y) \sim N_{mm}(\log\{D(X)\},0.2^2) \]
    where $D(X)=\exp\begin{pmatrix}
        1 & \zeta(X) \\ \zeta(X) & 1
    \end{pmatrix}$ and $\zeta(X)= \sin\Big(4\pi\beta^\top X \cdot (2\beta^\top X-1)\Big)$.

    \textbf{Setting II-2}:
    \[\log(Y) \sim N_{mm}(\log\{D(X)\},0.2^2) \]
    where $D(X)=\exp\begin{pmatrix}
        1 & \zeta_1(X) & \zeta_2(X) \\ 
        \zeta_1(X) & 1 & \zeta_1(X) \\
        \zeta_2(X) & \zeta_1(X) & 1  
    \end{pmatrix}$ and $\zeta_1(X) = x_1/\left(1+|x_2|^\frac{1}{2}\right)$, 
    $\zeta_2(X) = \sin\left(x_3^2\right) + \exp(x_4^2)$.
    The two settings correspond to cases where the structural dimension is $d=1$ and $d=2$, respectively.

    The results in Tables \ref{SPD_1} and \ref{SPD_2} show that our method outperforms WIRE in terms of the accuracy of estimated sufficient predictors for SPD responses across different metrics, demonstrating its robustness.
    \begin{table}[htbp]
        \centering
        \renewcommand{\arraystretch}{0.6} 
        \caption{Distance correlations for models with SPD responses and Log-Cholesky metric.}
        \label{SPD_1}
        \scalebox{0.9}{
        \begin{tabular}{@{\hspace{25pt}}c@{\hspace{25pt}}c@{\hspace{10pt}}c@{\hspace{25pt}}c@{\hspace{25pt}}c@{\hspace{25pt}}}
            \toprule
            Models & $(p,n)$ &  & FCCov-Net & WIRE \\
            \midrule
            & (10, 1000) &  & \textbf{0.832(0.023)} & 0.645(0.002)  \\
            & (10, 2000) &  & \textbf{0.905(0.026)} & 0.658(0.001) \\
            \textbf{Setting II-1}& (10, 5000) &  & \textbf{0.979(0.000)} & 0.673(0.000) \\
            & (30, 1000) &  & \textbf{0.626(0.022)} & 0.572(0.001)  \\
            & (30, 2000) &  & \textbf{0.759(0.017)} & 0.604(0.001) \\
            & (30, 5000) &  & \textbf{0.835(0.072)} & 0.643(0.000) \\
            \\
            & (10, 1000) &  & \textbf{0.897(0.000)} & 0.893(0.000)  \\
            & (10, 2000) &  & \textbf{0.897(0.000)} & 0.890(0.000) \\
            \textbf{Setting II-2}& (10, 5000) &  & \textbf{0.913(0.000)} & 0.893(0.000) \\
            & (30, 1000) &  & 0.842(0.025) & \textbf{0.874(0.000)}  \\
            & (30, 2000) &  & \textbf{0.881(0.000)} & 0.876(0.000) \\
            & (30, 5000) &  & \textbf{0.891(0.000)} & 0.876(0.000) \\
            \bottomrule
          \end{tabular}
        }
    \end{table}

    \begin{table}[htbp]
        \centering
        \renewcommand{\arraystretch}{0.6} 
        \caption{Distance correlations for models with SPD responses and affine invariant metric.}
        \label{SPD_2}
        \scalebox{0.9}{
        \begin{tabular}{@{\hspace{25pt}}c@{\hspace{25pt}}c@{\hspace{10pt}}c@{\hspace{25pt}}c@{\hspace{25pt}}c@{\hspace{25pt}}}
            \toprule
            Models & $(p,n)$ &  & FCCov-Net & WIRE \\
            \midrule
            & (10, 1000) & & \textbf{0.804(0.026)} & 0.645(0.001) \\
            & (10, 2000) &  & \textbf{0.948(0.001)} & 0.656(0.001)  \\
            \textbf{Setting II-1}& (10, 5000) &  & \textbf{0.982(0.000)} & 0.666(0.000)  \\
            & (30, 1000) &  & \textbf{0.644(0.017)} & 0.622(0.002)  \\
            & (30, 2000) &  & \textbf{0.809(0.001)} & 0.643(0.001)  \\
            & (30, 5000) &  & \textbf{0.949(0.000)} & 0.659(0.000)  \\
            \\
            & (10, 1000) &  & \textbf{0.929(0.000)} & 0.895(0.000) \\
            & (10, 2000) &  & \textbf{0.934(0.000)} & 0.894(0.000) \\
            \textbf{Setting II-2}& (10, 5000) &  & \textbf{0.938(0.000)} & 0.894(0.000)  \\
            & (30, 1000) &  & \textbf{0.896(0.000)} & 0.871(0.000) \\
            & (30, 2000) &  & \textbf{0.930(0.000)} & 0.894(0.000)  \\
            & (30, 5000) &  & \textbf{0.925(0.000)} & 0.878(0.000)  \\
            \bottomrule
          \end{tabular}
        }
    \end{table}

    \section{Real Data}\label{sec:real_data}

    To evaluate the effectiveness of our proposed FCCov-Net method in practical applications, we applied our method to the JAFFE dataset. We randomly divided the original dataset into a training set comprising 170 samples and a test set containing 43 samples.

    

    \begin{figure}[htbp]
        \centering
        \includegraphics[width=0.68\textwidth]{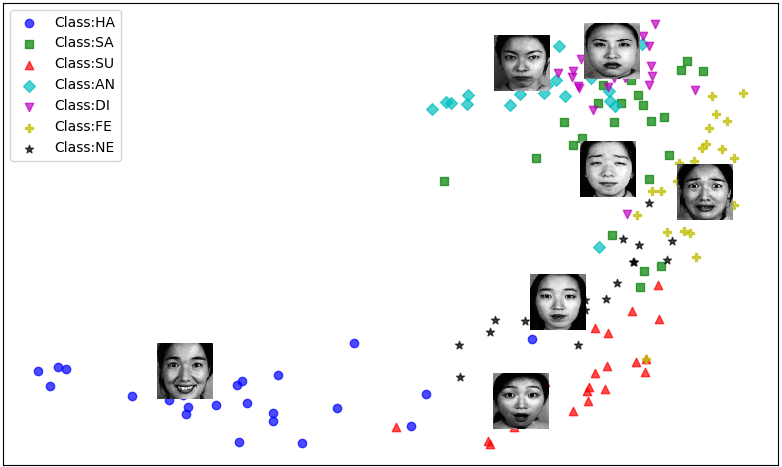}
        \caption{Scatter plots of the first two sufficient predictors estimated by FCCov-Net.}
        \label{2d_feature}
    \end{figure}
    Figure \ref{2d_feature} presents the scatter plots of the first two sufficient
    predictors estimated by FCCov-Net with Hellinger distance based on the training data.
The scatter plot depicts a vertical gradient of facial expressions, ranging from positive at the bottom, neutral in the middle, to negative at the top.
    Below the neutral points, happiness images are located on the left and surprise ones on the right. Above the neutral points, anger images appear on the left and fear ones on the right. Disgust images are clustered at the top-center, while sadness ones are positioned at the bottom-center. We select some representative points and show their facial images.

    Subsequently, we compared FCCov-Net and WIRE using Hellinger distance (HD) and Total Variation (TV) distance as metrics for responses being probability vectors.  Other than the emotion distribution labels, this dataset also have expression category labels. These labels indicate the specific expression category to which each image belongs. They are utilized as label data in classification tasks.
    We employ $\hat{\mf}(X)$ to perform Fr\'echet regression with distributional responses and classification with categorical labels, where $\hat{\mf} \in \mathbb{R}^d$ represents nonlinear SDR based on the training set. 
    \begin{table}[htbp]
        \centering
        \caption{The percentage of correct classifications for the testing group based on $\hat{f}(\cdot)$}
        \label{tab:jaffe_acc}
        \begin{tabular}{@{}lccccccccccc@{}}
        \toprule
        & \multicolumn{3}{c}{\( d = 1 \)} & \multicolumn{3}{c}{\( d = 2 \)} & \multicolumn{3}{c}{\( d = 3 \)} \\
        \cmidrule(r){2-4} \cmidrule(lr){5-7} \cmidrule(lr){8-10}
        Algorithm & LR & RF & SVM & LR & RF & SVM & LR & RF & SVM \\
        \midrule
        FCCovNet-HD & 0.35 & 0.46 & 0.51 & 0.53 & 0.58 & 0.67 & 0.72 & 0.77 & 0.81  \\
        WIRE-HD & 0.27 & 0.35 & 0.10 & 0.29 & 0.58 & 0.14 & 0.53 & 0.63 & 0.28  \\
        FCCovNet-TV & 0.4 & 0.38 & 0.44 & 0.52 & 0.54 & 0.58 & 0.70 & 0.73 & 0.73 \\
        WIRE-TV & 0.18 & 0.34 & 0.12 & 0.37 & 0.58 & 0.17 & 0.45 & 0.70 & 0.20 \\
        \bottomrule
        \end{tabular}
    \end{table}

    For classification, we employ three algorithms: logistic regression (LR), random forests (RF), and support vector machines (SVM). We then calculate the accuracy of these algorithms on the test set, with the results presented in Table \ref{tab:jaffe_acc}. As illustrated in the table, FCCov-Net demonstrates higher classification accuracy than WIRE across different $d$. 
    
    In addition to expression category labels, emotion distributions are crucial for reflecting the effects of nonlinear SDR methods. Consequently, leveraging local Fr\'echet regression with Hellinger distance as described in \citet{petersen2019frechet}, we utilize $d$-dimensional features to predict the corresponding label distributions. Various measures of similarity between predicted and true label distributions are summarized in Table \ref{tab:measures}. The results for Fr\'echet regression based on $\hat\mf(X)$, using different distribution distance metrics for test data are presented in Table \ref{tab:LDL}. 
    Both Table \ref{tab:jaffe_acc} and \ref{tab:LDL} demonstrate that our method obtains more accurate sufficient predictors while preserving sufficient information and remains robust across different distance metrics.
    
\begin{figure}[ht]
        \centering
        \subfloat[Front view]{%
            \includegraphics[width=0.4\textwidth]{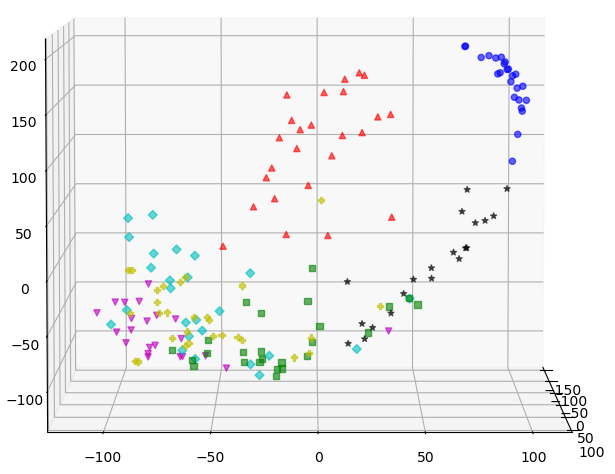}
            \label{fig:sub1}
        }
        \subfloat[Top view]{%
            \includegraphics[width=0.35\textwidth]{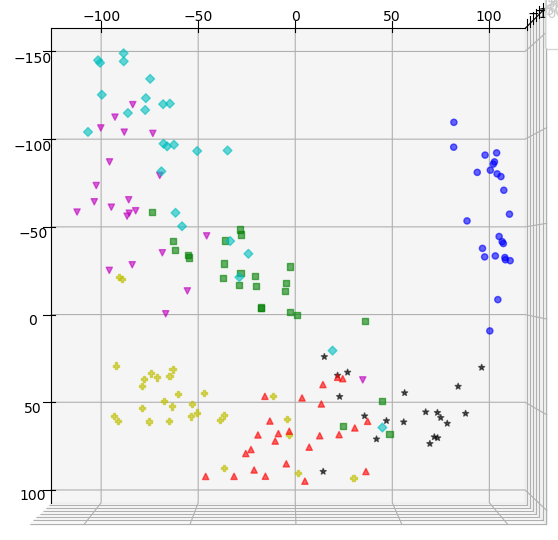}
            \label{fig:sub3}
        }
        \subfloat{%
            \raisebox{1cm}{\includegraphics[width=0.2\textwidth]{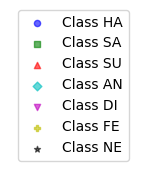}}
            \label{fig:colorbar}
        }
        \caption{Scatter plots of the first three sufficient predictors estimated by FCCov-Net.}
        \label{fig:3d_feature}
    \end{figure}

   From the results of Table \ref{tab:LDL}, we see that the test error with $d=3$ is the smallest, which indicates that we can select $d$ as $3$. In addition, the two-dimensional scatter plots exhibit overlapping expressions. In contrast, three-dimensional scatter plots viewed from the top as shown in Figure \ref{fig:3d_feature} clearly distinguish different emotion categories.

    \begin{table}[ht]
        \centering
        \caption{Distance metrics for Fr\'echet regression with responses beging probability vectors.}
        \label{tab:measures}
        \begin{tabular}{|c|c|}
            \hline Name & Formula \\
            \hline \hline Kullback-Leibler(K-L) &$\operatorname{Dis}_1=\sum_{j=1}^C P_j \ln \frac{P_j}{Q_j}$ \\
            Euclidean & $\operatorname{Dis}_2=\sqrt{\sum_{j=1}^C\left(P_j-Q_j\right)^2}$ \\
            Sørensen & $\operatorname{Dis}_3=\frac{\sum_{j=1}^C\left|P_j-Q_j\right|}{\sum_{j=1}^C\left(P_j+Q_j\right)}$ \\
            Squared $X^2$ & $\operatorname{Dis}_4=\sum_{j=1}^C \frac{\left(P_j-Q_j\right)^2}{P_j+Q_j}$ \\
            \hline
        \end{tabular}
    \end{table}

    \begin{table}[htbp]
        \centering
        \caption{Distances between the predicted distributions 
        and the true distributions.}
        \label{tab:LDL}
        \resizebox{\columnwidth}{!}{
        \begin{tabular}{@{}lcccccccccccc@{}}
        \toprule
        & \multicolumn{4}{c}{\( d = 1 \)} & \multicolumn{4}{c}{\( d = 2 \)} & \multicolumn{4}{c}{\( d = 3 \)} \\
        \cmidrule(r){2-5} \cmidrule(lr){6-9} \cmidrule(l){10-13}
        Algorithm & $\operatorname{Dis}_1$ & $\operatorname{Dis}_2$ & $\operatorname{Dis}_3$ & $\operatorname{Dis}_4$
        & $\operatorname{Dis}_1$ & $\operatorname{Dis}_2$ & $\operatorname{Dis}_3$ & $\operatorname{Dis}_4$
        & $\operatorname{Dis}_1$ & $\operatorname{Dis}_2$ & $\operatorname{Dis}_3$ & $\operatorname{Dis}_4$ \\
        \midrule
        FCCovNet-HD & 0.0968 & 0.0321 & 0.0918 & 1.3429 & 0.0779 & 0.0226 & 0.0761 & 0.9542 & 0.0574 & 0.0113 & 0.0571 & 0.4880 \\
        WIRE-HD & 0.1082 & 0.0347 & 0.1037 & 1.4781 & 0.0860 & 0.0233 & 0.0851 & 0.9976 & 0.0718 & 0.0162 & 0.0690 & 0.6999 \\
        FCCovNet-TV & 0.1035 & 0.0367 & 0.0978 & 1.5513 & 0.0868 & 0.0304 & 0.0838 & 1.2437 & 0.0644 & 0.0160 & 0.0635 & 0.6782 \\
        WIRE-TV & 0.1119 & 0.0369 & 0.1077 & 1.5855 & 0.0913 & 0.0269 & 0.0897 & 1.1415 & 0.0871 & 0.0233 & 0.0865 & 1.0064 \\
        \bottomrule
        \end{tabular}
        }
    \end{table}
    
\section{Conclusion}\label{sec:summary}

    In this paper, we propose a novel method for nonlinear SDR
    with general metric space valued responses based on a new measure 
    to characterize statistical dependence. We conduct a systematic analysis of the theoretical properties of our proposal, including ubiasedness in the population level, feasibility from the computational perspective, and nonasymptotic property of the estimator based on neural networks. 
    
   In light of the significant advancements in the research on measuring statistical dependence in recent years  (e.g., \citet{JMLR:v23:20-682, liu2022model, tong2023model}), the integration of effective dependence measures into our framework also warrants further investigation. Moreover, adopting advanced neural architectures like transformers could further enhance the performance of nonlinear SDR for complex dependent data structures. We leave this for future research.

\bibliographystyle{apalike}

\bibliography{jasaref}

\end{document}